\documentclass[fleqn,10pt]{wlscirep}
\usepackage{siunitx}
\usepackage{graphicx}

\title{Branching embedding: A heuristic dimensionality reduction algorithm based on hierarchical clustering}

\author[1,*]{Makito Oku}
\affil[1]{Institute of Natural Medicine, University of Toyama, Japan}
\affil[*]{oku@inm.u-toyama.ac.jp}


\begin{abstract}
This paper proposes a new dimensionality reduction algorithm named branching embedding (BE). It converts a dendrogram to a two-dimensional scatter plot, and visualizes the inherent structures of the original high-dimensional data. Since the conversion part is not computationally demanding, the BE algorithm would be beneficial for the case where hierarchical clustering is already performed. Numerical experiments revealed that the outputs of the algorithm moderately preserve the original hierarchical structures.

\end{abstract}
\begin{document}

\flushbottom
\maketitle

\thispagestyle{empty}

\section{Introduction}

Dimensionality reduction is the process of converting high-dimensional data to low-dimensional data. Its outcome is generally presented as a two- or three-dimensional scatter plot. Dimensionality reduction algorithms include principal component analysis (PCA)\cite{jolliffe1986}, nonnegative matrix factorization (NMF)\cite{lee1999}, multidimensional scaling (MDS)\cite{kruskal1964}, Isomap\cite{tenenbaum2000}, locally linear embedding (LLE)\cite{roweis2000}, Laplacian eigenmaps\cite{belkin2003}, diffusion maps\cite{coifman2006}, t-distributed stochastic neighbor embedding (t-SNE)\cite{maaten2008}, uniform manifold approximation and projection (UMAP)\cite{mcinnes2018umap}, and so on. 

On the other hand, agglomerative hierarchical clustering is widely used in various research fields, especially in bioinformatics. Since non-agglomerative (divisive) hierarchical clustering is rarely used, the adjective 'agglomerative' is omitted henceforth. Although hierarchical clustering requires relatively large computational costs, it is commonly used with heatmap plots because dendrograms are suitable for being shown on a side of a heatmap plot.

Since scatter plots are simpler than dendrograms, conversion from a dendrogram to a scatter plot would be helpful to interpret the outcomes of hierarchical clustering. A related idea is to regard a dendrogram as a tree graph and perform an existing graph layout algorithm. However, the outcome tends to be messy due to the presence of edges and non-leaf vertices.

\begin{figure}[ht]
\centering
\includegraphics[width=0.8\linewidth]{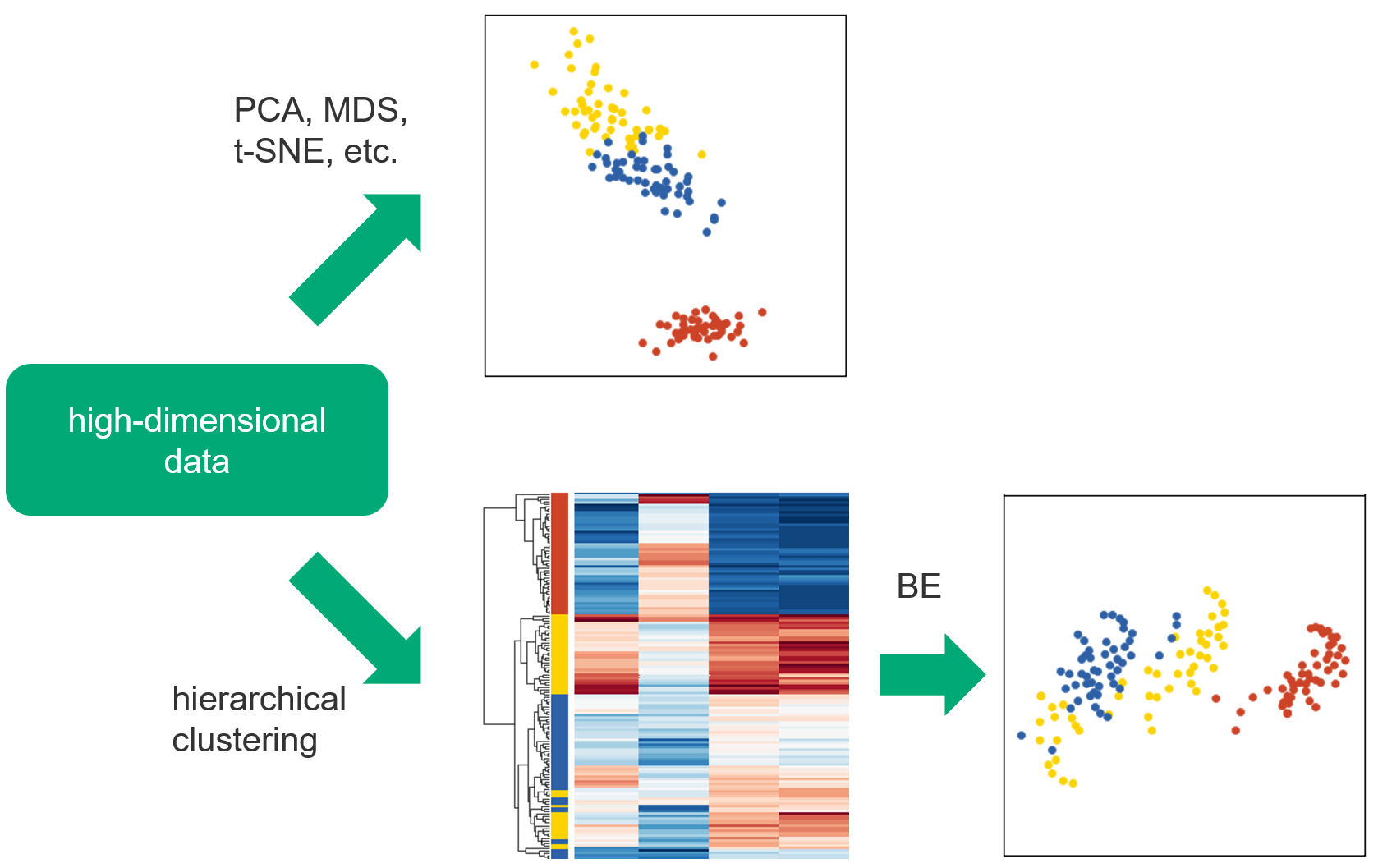}
\caption{Schematic illustration of the problem setting considered in this study. The subplots were generated using the iris dataset.}
\label{fig01}
\end{figure}

In this paper, I propose an algorithm named branching embedding (BE) for converting a dendrogram to a two-dimensional scatter plot (Fig.~\ref{fig01}). The goal is to assign a position to each leaf node while retaining the original hierarchical structure as much as possible. Since its computational cost is not high, a potential target of the BE algorithm is the case where hierarchical clustering, which is computationally demanding, is already performed. 

The goodness of embedding is measured by the discrepancy between the original dendrogram and the dendrogram re-calculated from the two-dimensional output. Although the BE algorithm does not guarantee the perfect match, numerical experiments show that the two dendrograms are moderately correlated in many cases of artificial or real data.

\section{Basic properties of dendrograms}

This section provides a brief overview of the basic properties of dendrograms. \textit{Cophenetic dissimilarity} of two leaf nodes of a dendrogram is the height of the closest common ancestor of them. When a dendrogram is monotonic, cophenetic dissimilarity satisfies the ultrametric inequality, which is a stronger condition than the triangle inequality, and is called \textit{cophenetic distance}. Two dendrograms are called \textit{equivalent} if they have the same cophenetic matrix. 

We define \textit{the degree of kinship} of two leaf nodes of a dendrogram as the minimum number of edges required to move from one to the other. It satisfies the ultrametric inequality even if a dendrogram is non-monotonic. A kinship matrix is defined as a matrix showing the degrees of kinship of all leaf node pairs. We call two dendrograms \textit{isomorphic} if they have the same kinship matrix. 

In this paper, the original dendrogram and that obtained from the two-dimensional outputs (the converted dendrogram) are compared. The latter one is assumed to be calculated using the Euclidean distance and the same linkage method to the original one. Both dendrograms are assumed to be monotonic.

It is always possible to make the original and converted dendrograms equivalent if the single linkage method is used. A constructive proof is to put the leaf nodes along a straight line in the order appearing in the original dendrogram, and separate each pair of adjacent points according to the cophenetic distance between them. The resulting path graph can be bent on a two-dimensional plane as long as the hierarchical structure is preserved. However, the single linkage method frequently causes the chaining problem in bioinformatics researches, and thus other linkage methods are preferred. 

Unfortunately, it is not always possible to make the original and converted dendrograms equivalent if the average or complete linkage method is used. A counterexample is an original dendrogram of four leaf nodes 1, 2, 3, and 4 with $d(1,2)=d(3,4)=1$ and $d(1,3)=d(1,4)=d(2,3)=d(2,4)=l>1$, where $d$ is the cophenetic distance. If the average linkage method is used and $l<(1+\sqrt{2})/2$, or if the complete linkage method is used and $l<\sqrt{2}$, no two-dimensional point set is able to reproduce the same cophenetic matrix.

Alternatively, we can always make the two dendrograms isomorphic if the average or complete linkage method is used. A constructive proof begins with allocating distinct two-dimensional planes to each of the leaf nodes. Subclusters are merged in the same order to the original dendrogram. At each merge event, the two point sets belonging to the two subclusters being merged are put together on the same two-dimensional plane. Their distance is set to be sufficiently large in order to guarantee that the two point sets are merged at last if hierarchical clustering is performed against the points on the plane. However, to the best of the author's knowledge, an efficient algorithm for achieving a sufficiently compact output has not been developed.

The properties of the distance matrix associated with the two-dimensional output is out of the scope of this paper. However, it is important to note that the distance matrix cannot be the same to the original cophenetic matrix if the original dendrogram is monotonic and has four or more distinct leaf nodes. The proof is not difficult and thus omitted.

\begin{figure}[ht]
\centering
\includegraphics[width=0.7\linewidth]{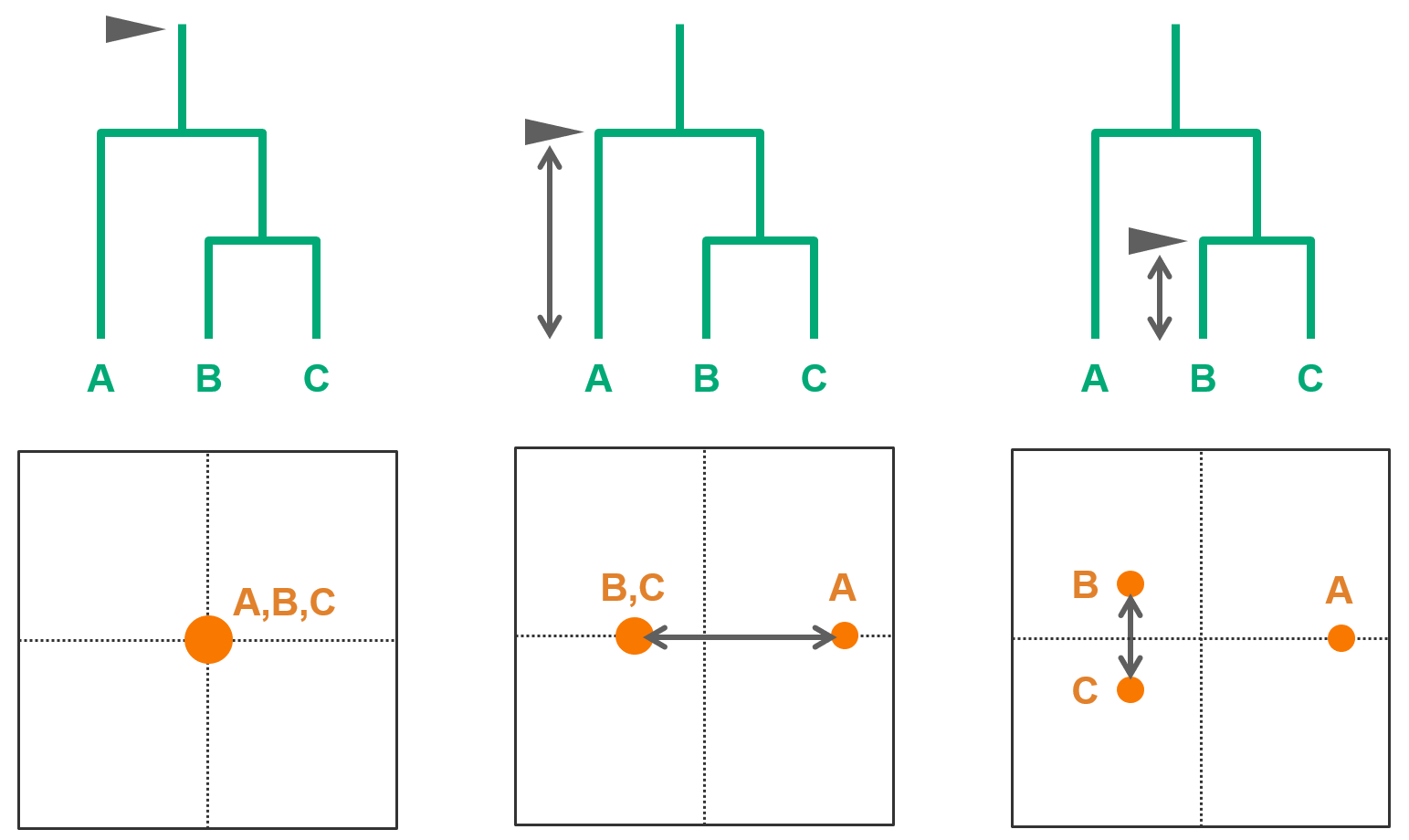}
\caption{Schematic illustration of the BE algorithm.}
\label{fig02}
\end{figure}

\begin{figure}[ht]
\centering
\includegraphics[width=0.7\linewidth]{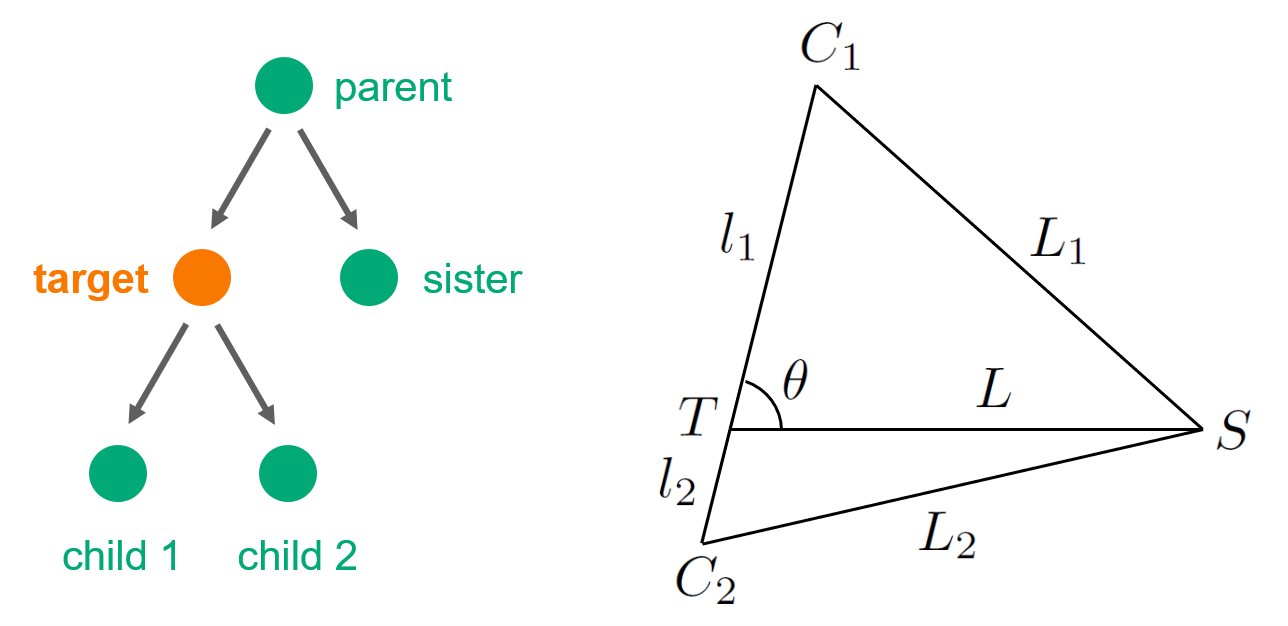}
\caption{Schematic illustration of the division process. (left) Relationship between the target cluster and its parent, sister, and child clusters. (right) Two-dimensional positions of the target ($T$), sister ($S$), child 1 ($C_1$) and child 2 ($C_2$) clusters.}
\label{fig03}
\end{figure}

\section{The BE algorithm}

Figure~\ref{fig02} shows a schematic illustration of the BE algorithm. It begins with the root of a dendrogram, where all leaf nodes are put into a single cluster. We assign (0,0) to it. The first cluster splits into two subclusters at the first branching, and the corresponding positions move in opposite directions. We set the distance between them to the height of the branching event in the dendrogram. In order to keep the center of mass unchanged during the division process, we make the travel distance of each subcluster inversely proportional to its size. The subclusters are repeatedly divided according to the dendrogram, and finally all leaf nodes are isolated and assigned two-dimensional coordinates.

The division angle needs careful considerations. The first branching can be in any direction, but the directions of subsequent divisions affect the goodness of embedding. A simple method is to allocate a random angle to each branching. Another naive approach is to alternate between vertical and horizontal divisions according to the lineage history. A generalization of this method is to fix the angle $\theta$ in Fig.~\ref{fig03}. It is the angle between two line segments, one from the target cluster to sister cluster, and the other from the target cluster to child 1 cluster. When $\theta$ is small (for example, $\theta\simeq\ang{15}$), the goodness of embedding is often improved if child 1 and child 2 are swapped when the former is bigger, in order to guarantee that the larger child cluster is always pushed away from the sister cluster. Another approach is to set $\theta=\cos^{-1}((l_1-l_2)/2L)$ so that $L_1=L_2$ is satisfied, which is called the 'even' method henceforth.

The computational complexity of the BE algorithm is $\mathrm{O}(n)$ excluding the hierarchical clustering part, where $n$ denotes the number of leaf nodes. It is easy to implement, and a python source code is made available in the GitHub repository under MIT license (\url{http://github.com/okumakito/branch-embed}). 

\section{Experimental results}

The performance of different angle determination strategies of the BE algorithm was initially compared using random matrices. Each matrix had 100 rows and 5 columns, and the element values were independently sampled from the standard normal distribution. The goodness of embedding was measured by two statistics: (1) the correlation coefficient between the original and converted cophenetic matrices $r_c$ and (2) the correlation coefficient between the original and converted kinship matrices $r_k$. Since the cophenetic and kinship matrices are symmetric and their diagonal elements are 0 by definition, only the elements above the main diagonal were considered. 

The original dissimilarity matrix were calculated using either the Euclidean distance or the sign-reversed correlation coefficient added by 1. It is important to note that the latter does not always satisfy the triangle inequality and should not be called distance or metric. For the case of the Euclidean distance, the single, complete, average, and Ward linkage methods were used. For the case of the correlation similarity, the single, complete, and average linkage methods were investigated.

Tables~\ref{table01} and \ref{table02} show the results of $r_c$ and $r_k$, respectively. The random angle method, the perpendicular division method ($\theta=\ang{90}$), and the even method were not optimal in any case. The fixed angle method turned out to be a good choice, and the best angle depended on the clustering condition. The $\ang{15}$ angle was optimal in many cases. A possible reason is that the encounter rate of different offspring groups are suppressed due to the formation of spiral-like patterns. For the case of the Ward's method, the $\ang{45}$ and \ang{60} angles were the best. For the case of the Euclidean distance and the single linkage method, the $\ang{75}$ angle achieved the largest $r_c$. However, the corresponding value of $r_k$ was very low. For the case of the correlation similarity and the single linkage method, the $\ang{0}$ angle was optimal. However, its output is one-dimensional and may not be suitable for visualization purposes. The performance also depended on the size of the random matrices.

\begin{table}[t]
\centering
\caption{\label{table01}Average correlation coefficients between the original and converted cophenetic matrices. A total of 1000 random matrices with 100 rows and 5 columns were used.}
\begin{tabular}{l|rrrrrrrrr}
\toprule
clustering condition & random & \ang{0} & \ang{15} & \ang{30} & \ang{45} & \ang{60} & \ang{75} & \ang{90} & even \\
\midrule
Euclidean, single & 0.40 & 0.07 & 0.41 & 0.43 & 0.38 & 0.32 & \textsf{\textbf{0.46}} & 0.23 & 0.31 \\
Euclidean, complete & 0.45 & 0.28 & 0.44 & \textsf{\textbf{0.51}} & 0.48 & 0.47 & 0.47 & 0.46 & 0.48 \\
Euclidean, average & 0.29 & 0.17 & \textsf{\textbf{0.43}} & 0.31 & 0.27 & 0.25 & 0.25 & 0.25 & 0.26 \\
Euclidean, Ward & 0.61 & 0.40 & 0.52 & 0.66 & \textsf{\textbf{0.71}} & \textsf{\textbf{0.71}} & 0.69 & 0.69 & 0.69 \\
correlation, single & 0.12 & \textsf{\textbf{0.15}} & \textsf{\textbf{0.15}} & 0.12 & 0.11 & 0.11 & 0.11 & 0.11 & 0.12 \\
correlation, complete & 0.30 & 0.29 & \textsf{\textbf{0.40}} & 0.29 & 0.28 & 0.29 & 0.31 & 0.30 & 0.31 \\
correlation, average & 0.37 & 0.36 & \textsf{\textbf{0.53}} & 0.43 & 0.35 & 0.34 & 0.35 & 0.35 & 0.36 \\
\bottomrule
\end{tabular}
\end{table}

\begin{table}[t]
\centering
\caption{\label{table02}Average correlation coefficients between the original and converted kinship matrices. A total of 1000 random matrices with 100 rows and 5 columns were used.}
\begin{tabular}{l|rrrrrrrrr}
\toprule
clustering condition & random & \ang{0} & \ang{15} & \ang{30} & \ang{45} & \ang{60} & \ang{75} & \ang{90} & even \\
\midrule
Euclidean, single & 0.05 & 0.15 & \textsf{\textbf{0.16}} & 0.06 & 0.00 & -0.02 & -0.01 & -0.03 & 0.02 \\
Euclidean, complete & 0.41 & 0.21 & 0.37 & \textsf{\textbf{0.46}} & 0.44 & 0.44 & 0.44 & 0.43 & 0.44 \\
Euclidean, average & 0.22 & 0.12 & \textsf{\textbf{0.26}} & 0.22 & 0.20 & 0.19 & 0.19 & 0.20 & 0.20 \\
Euclidean, Ward & 0.59 & 0.33 & 0.49 & 0.63 & 0.70 & \textsf{\textbf{0.73}} & 0.72 & \textsf{\textbf{0.73}} & 0.71 \\
correlation, single & 0.09 & \textsf{\textbf{0.18}} & 0.14 & 0.10 & 0.08 & 0.08 & 0.08 & 0.08 & 0.10 \\
correlation, complete & 0.16 & 0.18 & \textsf{\textbf{0.27}} & 0.20 & 0.15 & 0.14 & 0.14 & 0.15 & 0.16 \\
correlation, average & 0.23 & 0.25 & \textsf{\textbf{0.36}} & 0.29 & 0.23 & 0.21 & 0.21 & 0.21 & 0.23 \\
\bottomrule
\end{tabular}
\end{table}

\begin{figure}[ht!]
\centering
\includegraphics[width=1.0\linewidth]{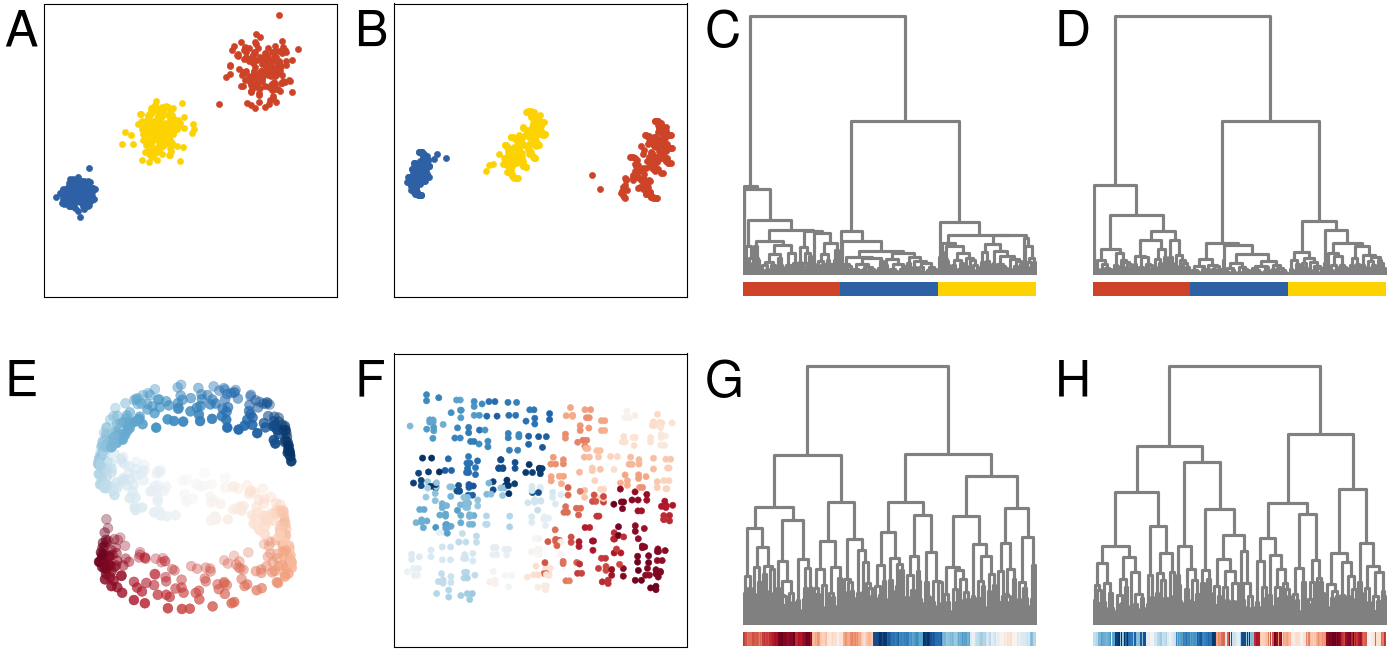}
\caption{Applications of the BE algorithm to two artificial datasets: a blob dataset (A--D) and a S-shaped dataset (E--H). (A and E) the original data points. (B and F) the outputs of the BE algorithm. (C and G) the original dendrograms. (D and H) the converted dendrograms.}
\label{fig04}
\end{figure}

The BE algorithm was then applied to two artificial datasets. Both were generated using the scikit-learn library in python. The first one is a blob dataset as shown in Fig.~\ref{fig04}A. A total of 500 points were randomly sampled from three two-dimensional isotropic normal distributions with standard deviations 0.5, 0.8, and 1. The center positions of the three normal distributions were randomly determined. The numbers of points belonging to the three blobs were almost balanced. Figure~\ref{fig04}C shows the original dendrogram calculated using the Euclidean distance and the average linkage method. Figure~\ref{fig04}B shows the output of the BE algorithm with $\theta=\ang{15}$. The three clusters were successfully separated, and the order relation of the cluster width was preserved. Figure\ref{fig04}D shows the converted dendrogram, which was strongly correlated with the original one ($r_c=0.997$ and $r_k=0.714$).

The second example is a S-shaped dataset as shown in Fig.~\ref{fig04}E. A total of 500 points were uniformly sampled from the S-shaped two-dimensional manifold in the three-dimensional Euclidean space. Figure~\ref{fig04}G shows the original dendrogram calculated using the Euclidean distance and the average linkage method. Figure~\ref{fig04}F shows the output of the BE algorithm with $\theta=\ang{90}$. It failed to unfold the S-shaped sheet completely, but local structures were well preserved. Figure\ref{fig04}H shows the converted dendrogram, which was moderately correlated with the original one ($r_c=0.696$ and $r_k=0.634$).

\begin{figure}[ht!]
\centering
\includegraphics[width=0.75\linewidth]{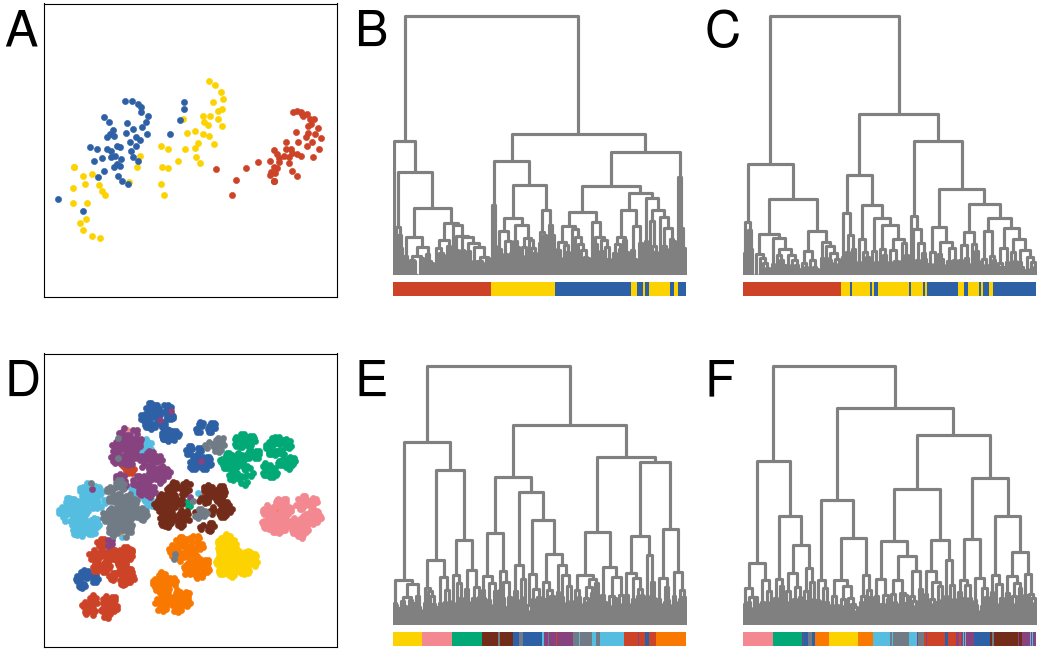}
\caption{Applications of the BE algorithm to two real datasets: the iris dataset (A--C) and the handwritten digit dataset (D--F). (A and D) the outputs of the BE algorithm. (B and E) the original dendrograms. (C and F) the converted dendrograms.}
\label{fig05}
\end{figure}

The BE algorithm was also applied to two real datasets. Both were commonly used as benchmark problems in machine learning researches and were loaded using the scikit-learn library. The first one is the iris dataset, which has 150 observations, 4 variables, and 3 classes. The sizes of the three classes were all 50. The raw values were rescaled between 0 and 1 for each variable. Figure~\ref{fig05}B shows the original dendrogram calculated using the Euclidean distance and the average linkage method. Figure~\ref{fig05}A shows the output of the BE algorithm with $\theta=\ang{15}$. One cluster was successfully separated from the others, but the remaining two clusters were slightly overlapping. Figure\ref{fig05}C shows the converted dendrogram. Although it looks considerably different from the original dendrogram, the goodness of embedding measures were large ($r_c=0.967$ and $r_k=0.628$).

The second real-data example is the handwritten digit dataset, which has 1797 observations, 64 variables, and 10 classes. The sizes of the 10 classes were almost balanced. Figure~\ref{fig05}E shows the original dendrogram calculated using the Euclidean distance and the Ward's method. Figure~\ref{fig05}D shows the output of the BE algorithm with $\theta=\ang{60}$. The 10 classes were well separated from each other, but some of them were further split into smaller subclusters. This does not necessarily indicate the existence of corresponding subclusters in the original high-dimensional space. In fact, the emergence of unexpected subclusters is commonly observed when the BE algorithm is applied to a dendrogram calculated using the Ward's method. Figure\ref{fig05}F shows the converted dendrogram, which was largely correlated with the original one ($r_c=0.742$ and $r_k=0.629$). It was calculated using the average linkage method instead of the Ward's method because when the Ward's method was used in both cases, the appearance of the converted dendrogram was considerably different from the original one. 

\section{Conclusions}

This study proposed the BE algorithm, a novel dimensionality reduction method based on hierarchical clustering. Three angle determination methods (random angle, fixed angle, and the even method) were considered and their performance was compared using the random matrices. The effectiveness of the BE algorithm was also demonstrated using the two artificial and two real datasets. The original hierarchical structures were moderately preserved in the sense that a similar dendrogram to the original one could be reproduced from the two-dimensional points. The BE algorithm would be suitable for converting a precomputed dendrogram to a two-dimensional scatter plot in order to interpret the results of hierarchical clustering.

%\bibliography{references}

\begin{thebibliography}{1}
\expandafter\ifx\csname url\endcsname\relax
  \def\url#1{\texttt{#1}}\fi
\expandafter\ifx\csname urlprefix\endcsname\relax\def\urlprefix{URL }\fi
\expandafter\ifx\csname doiprefix\endcsname\relax\def\doiprefix{DOI }\fi
\providecommand{\bibinfo}[2]{#2}
\providecommand{\eprint}[2][]{\url{#2}}

\bibitem{jolliffe1986}
\bibinfo{author}{Jolliffe, I.~T.}
\newblock \emph{\bibinfo{title}{Principal Component Analysis}}
  (\bibinfo{publisher}{Springer-Verlag, New York}, \bibinfo{year}{1986}),
  \bibinfo{edition}{2} edn.

\bibitem{lee1999}
\bibinfo{author}{Lee, D.~D.} \& \bibinfo{author}{Seung, H.~S.}
\newblock \bibinfo{journal}{\bibinfo{title}{Learning the parts of objects by
  non-negative matrix factorization}}.
\newblock {\emph{Nature}} \textbf{\bibinfo{volume}{401}},
  \bibinfo{pages}{788--791} (\bibinfo{year}{1999}).
\newblock \urlprefix\url{http://dx.doi.org/10.1038/44565}.

\bibitem{kruskal1964}
\bibinfo{author}{Kruskal, J.~B.}
\newblock \bibinfo{journal}{\bibinfo{title}{Multidimensional scaling by
  optimizing goodness of fit to a nonmetric hypothesis}}.
\newblock {\emph{Psychometrika}} \textbf{\bibinfo{volume}{29}},
  \bibinfo{pages}{1--27} (\bibinfo{year}{1964}).
\newblock \urlprefix\url{http://dx.doi.org/10.1007/BF02289565}.

\bibitem{tenenbaum2000}
\bibinfo{author}{Tenenbaum, J.~B.}, \bibinfo{author}{de~Silva, V.} \&
  \bibinfo{author}{Langford, J.~C.}
\newblock \bibinfo{journal}{\bibinfo{title}{A global geometric framework for
  nonlinear dimensionality reduction}}.
\newblock {\emph{Science}} \textbf{\bibinfo{volume}{290}},
  \bibinfo{pages}{2319--2323} (\bibinfo{year}{2000}).
\newblock \urlprefix\url{http://dx.doi.org/10.1126/science.290.5500.2319}.

\bibitem{roweis2000}
\bibinfo{author}{Roweis, S.~T.} \& \bibinfo{author}{Saul, L.~K.}
\newblock \bibinfo{journal}{\bibinfo{title}{Nonlinear dimensionality reduction
  by locally linear embedding}}.
\newblock {\emph{Science}} \textbf{\bibinfo{volume}{290}},
  \bibinfo{pages}{2323--2326} (\bibinfo{year}{2000}).
\newblock \urlprefix\url{http://dx.doi.org/10.1126/science.290.5500.2323}.

\bibitem{belkin2003}
\bibinfo{author}{Belkin, M.} \& \bibinfo{author}{Niyogi, P.}
\newblock \bibinfo{journal}{\bibinfo{title}{Laplacian eigenmaps for
  dimensionality reduction and data representation}}.
\newblock {\emph{Neural Comp.}} \textbf{\bibinfo{volume}{15}},
  \bibinfo{pages}{1373--1396} (\bibinfo{year}{2003}).
\newblock \urlprefix\url{http://dx.doi.org/10.1162/089976603321780317}.

\bibitem{coifman2006}
\bibinfo{author}{Coifman, R.~R.} \& \bibinfo{author}{Lafon, S.}
\newblock \bibinfo{journal}{\bibinfo{title}{Diffusion maps}}.
\newblock {\emph{Appl. Comp. Harm. Anal.}}
  \textbf{\bibinfo{volume}{21}}, \bibinfo{pages}{5--30} (\bibinfo{year}{2006}).
\newblock \urlprefix\url{http://dx.doi.org/10.1016/j.acha.2006.04.006}.

\bibitem{maaten2008}
\bibinfo{author}{van~der Maaten, L.} \& \bibinfo{author}{Hinton, G.}
\newblock \bibinfo{journal}{\bibinfo{title}{Visualizing data using {t-SNE}}}.
\newblock {\emph{J. Machine Learn. Res.}}
  \textbf{\bibinfo{volume}{9}}, \bibinfo{pages}{2579--2605}
  (\bibinfo{year}{2008}).
\newblock \urlprefix\url{http://www.jmlr.org/papers/v9/vandermaaten08a.html}.

\bibitem{mcinnes2018umap}
\bibinfo{author}{McInnes, L.} \& \bibinfo{author}{Healy, J.}
\newblock \bibinfo{journal}{\bibinfo{title}{{UMAP}: Uniform manifold
  approximation and projection for dimension reduction}}.
\newblock {\emph{arXiv:1802.03426}}  (\bibinfo{year}{2018}).
\newblock \urlprefix\url{https://arxiv.org/abs/1802.03426}.

\end{thebibliography}

\section*{Acknowledgments}

This work was supported by JSPS KAKENHI Grant Number JP15H05707.

\end{document}